# Perceptual Robust Hashing for Color Images with Canonical Correlation Analysis

Xinran Li[1], Chuan Qin[1], Zhenxing Qian[2], Heng Yao[1], and Xinpeng Zhang[2]

[1]School of Optical-Electrical and Computer Engineering, University of Shanghai for Science and Technology, Shanghai 200093, China
E-mail: ranusst@163.com, qin@usst.edu.cn, hyao@usst.edu.cn

[2]School of Computer Science, Fudan University, Shanghai 200433, China
E-mail: zxqian@fudan.edu.cn, zhangxinpeng@fudan.edu.cn

---

**Correspondence Address:**

Dr. Chuan Qin, Professor,

School of Optical-Electrical and Computer Engineering,

University of Shanghai for Science and Technology,

No. 516 Jungong Road, Shanghai 200093, China.

E-mail: qin@usst.edu.cn

TEL: 86-21-55272562

FAX: 86-21-55272982

# Perceptual Robust Hashing for Color Images with Canonical Correlation Analysis

Xinran Li, Chuan Qin, Zhenxing Qian, Heng Yao, and Xinpeng Zhang

**Abstract:** In this paper, a novel perceptual image hashing scheme for color images is proposed based on ring-ribbon quadtree and color vector angle. First, original image is subjected to normalization and Gaussian low-pass filtering to produce a secondary image, which is divided into a series of ring-ribbons with different radii and the same number of pixels. Then, both textural and color features are extracted locally and globally. Quadtree decomposition (QD) is applied on luminance values of the ring-ribbons to extract local textural features, and the gray level co-occurrence matrix (GLCM) is used to extract global textural features. Local color features of significant corner points on outer boundaries of ring-ribbons are extracted through color vector angles (CVA), and color low-order moments (CLMs) is utilized to extract global color features. Finally, two types of feature vectors are fused via canonical correlation analysis (CCA) to prodcue the final hash after scrambling. Compared with direct concatenation, the CCA feature fusion method improves classification performance, which better reflects overall correlation between two sets of feature vectors. Receiver operating characteristic (ROC) curve shows that our scheme has satisfactory performances with respect to robustness, discrimination and security, which can be effectively used in copy detection and content authentication.

## 1. Introduction

With the wide application of powerful image editing tools, digital images can be easily manipulated. Therefore, the authenticity and reliability of digital image contents are seriously threatened. In order to solve this problem, image hashing technique is introduced into the field of multimedia information security. Image hashing is also called image "fingerprint", which converts the original input image into a fixed-size string through a specific one-way mapping. Cryptographic hash functions are sensitive to all digital image manipulations. In other words, any slight changes in the input data will cause significant differences in the output results. Content retention manipulations, such as adding noise, scaling, rotation, compression, and filtering will cause the image input data to change, but the image content does not change substantially. Therefore, the cryptographic hash function is not suitable for multimedia authentication. Compared with the cryptographic hash function, image hash has more advantages. In the image authentication process, the image hash only depends on the perceived content of the image. Generally, a typical perceptual image hashing scheme include three main stages, including pre-processing, feature extraction and hash generation, as shown in Fig. 1. Three characteristics of an ideal image hashing scheme are as follows: (1) *Perceptual robustness*: images with similar visual perception after digital manipulations (such as image compression and enhancement) map to the same or similar "fingerprint" character strings. (2) *Discrimination*: also called anti-collision, the hashes generated from visually different images should be significantly different. It means that, there should be sufficiently large hash distances between visually different image pairs. (3) *Key-dependent security*: different secret keys produce completely different hashes, and it is difficult for the attacker to threaten the security of the scheme with brute force. In the case, when the secret key is unknown, no one can forge the correct hash sequence.

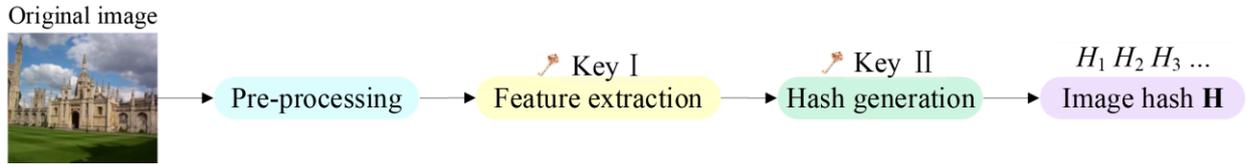

Fig. 1 The stages of a typical image hashing scheme.

Many earlier image hashing schemes focused on frequency domain transform, including discrete wavelet transform (DWT) [1], Radon transform (RT) [2], discrete Fourier transform (DFT) [3] and discrete cosine transform (DCT) [4], [5]. However, the classification ability of these schemes was insufficient, and the robustness to rotation was limited. Recently, a large number of image hashing schemes have been proposed through various strategies, such as non-negative matrix factorization (NMF) [6], Gaussian-Hermite moments (GHMs) [7], local binary pattern (LBP) [8], and tensor decomposition (TD) [9]. The scheme [10] presented an image hashing scheme to obtain the binary hash sequence, which conducted the block truncation coding (BTC) on the secondary image and then utilized center-symmetrical local binary pattern (CSLBP) to produce the final hash. However, this scheme had less performa nce in resisting rotation manipulation, and color features were not considered. Hamid *et al*. [11] used multi-scale filtering to extract image classification feature. Laplacian pyramids were generated with two different diameters filters to generate a robust hash. The scheme [12] extracted the ring-based entropy as image feature to obtain the hash, and these rings were divided from the normalized image. This scheme was robust to rotation manipulations, but ignored the color and textural features. In order to solve the geometric distortion problem, Liu *et al*. [13] proposed to use a secret key to divide the preprocessed image into several overlapping blocks randomly, and the hash codes were formed by invariant moment and low-frequency DCT coefficients. Vadlamudi *et al*. [14] performed scale-invariant feature transform (SIFT) feature points and DWT approximate coefficients to produce a unique hash. For most content-preserving manipulations, this scheme has satisfactory robustness, but its classification capability was not good enough. The scheme [15] utilized Canny operator to detect edge information, the dominant DCT coefficients of rich information blocks and the difference of rich and low information blocks are constructed into the final hash. In scheme [16], Davarzan *et al*.

used the inner product between the CSLBP feature vector and pseudo-random weight vector to extract perceptual features of images. This scheme has good robustness and anti-collision performance, but only small angle rotation can be resisted. Another scheme was introduced in [17], which was based on double cross pattern to extract textural perception and structural feature of the secondary image. Meanwhile, the blocks with significant structural features were selected, which contained the most corner points. The scheme [18] combined color and textural features with local binary mode and color vector angles (CVA) mode of Webers Law to construct image hash. A new quaternion image construction method was used to combine color and structural features in [19], which applied the quaternion Fourier-Merlin transform (QFMT) to calculate the feature hash. In addition to the above strategies, some researchers also designed image hashing schemes based on perceptual uniform descriptor (PUD) [20], color and luminance gradient feature [21], Watsons visual mode [22], color opponent component (COC) [23], speeded up robust features (SURF) [24], Delaunay triangulation and undirected graphs [25], and local linear embedding (LLE) [26].

Although the above-mentioned schemes can realize the basic functions of the hash schemes, there are still some shortcomings to be solved. For example, the perceptual robustness and anti-collision of these schemes cannot reach a well balance, and the classification and tampering detection performance of similar and different images are not satisfactory. The proposed hash scheme can effectively realize copy detection and content authentication by comparing the hash value of the original image and the suspicious image. The main contributions of the proposed scheme are listed as follows: (1) Textural and color features are integrated to present a robust image hashing scheme, which is based on ring-ribbon quadtree decomposition (QD) and CVA; (2) The pixels of each ring-ribbon maintain the same information after image rotation, hence, the scheme can resist image rotation manipulation; (3) Both local and global features are considered, and pixels of outside the inscribed circle are involved in feature extraction; (4) An effective hash generation strategy, which uses canonical correlation analysis (CCA) to integrate textural feature vectors with color feature vectors to improve the correlation between each kind of features.

The remaining parts of this paper are arranged as follows. The detailed steps of our perceptual image hashing scheme are described in Section 2. Section 3 presents the experimental results. Parameter selection and performance comparison are analyzed in Section 4. Section 5 makes the conclusions for the paper.

## 2. Proposed Scheme

In this section, our hashing scheme for color image based on local and global features is described detailedly. First, the original image $\mathbf{I}_o$ is normalized to generate the secondary image $\mathbf{I}$.

Then, our scheme divides the image **I** into a series of ring-ribbons **R**$_k$ to extract local textural feature and color feature, where $k = 1, 2, ..., N$, and $N$ is the number of ring-ribbons. Next, QD is applied to each ring-ribbon **R**$_k$ to produce the corresponding QD image and extract local textual feature. Meanwhile, significant corner points on the outer boundary of each ring-ribbon are selected to calculate CVA and extract local color feature. Furthermore, global textural feature and color feature are extracted through gray level co-occurrence matrix (GLCM) and color low-order moments (CLMs). Finally, the textural feature vectors and color feature vectors are fused via the extracted, local and global features to construct the final hash. The main symbols in our scheme and their meanings are listed in Table 1. Fig. 2 shows a schematic diagram of the proposed scheme.

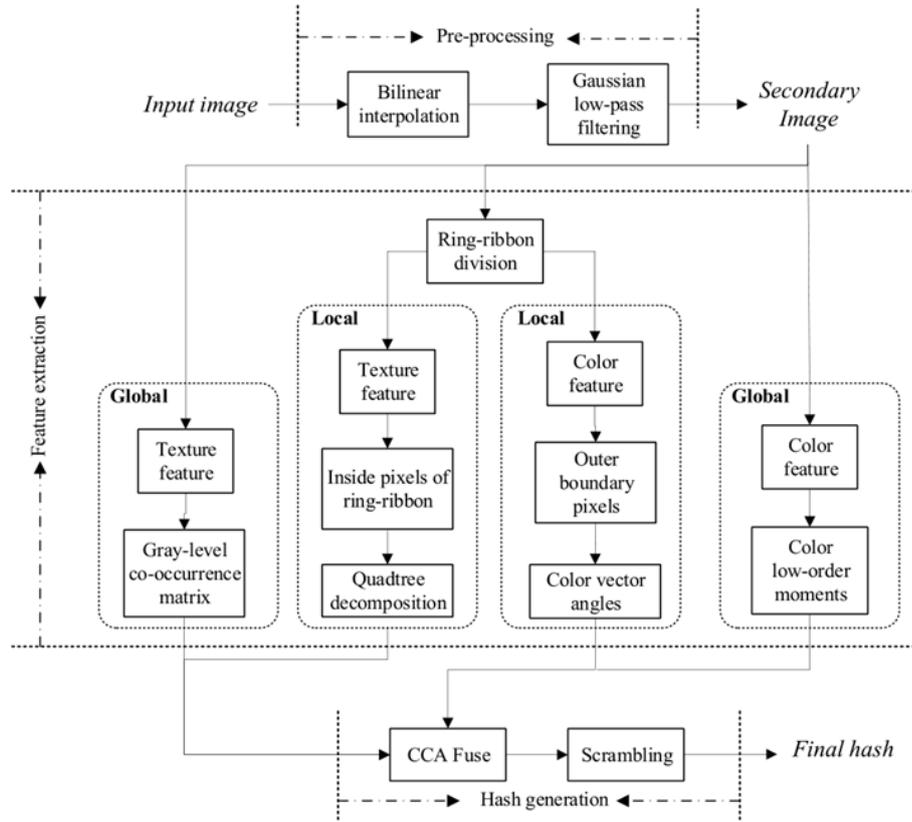

Fig. 2 A schematic diagram of the proposed image hashing scheme.

### 2.1 Pre-processing

The original image **I**$_o$ is resized to the standard resolution of $L \times L$ by bilinear interpolation, which can ensure that the hash lengths for different images are the same. Then, the image is smoothed with the Gaussian low-pass filter to produce the secondary image **I**. Gaussian low-pass filter uses a mask to process each pixel in the image to eliminate noise. The elements $E_G(u,v)$ of the mask are described by Eqs. (1)-(2).

Table I
THE MAIN SYMBOLS IN OUR ALGORITHM AND THEIR MEANINGS.

| Symbols | Meanings |
|---|---|
| $\mathbf{I}_o$ | The original image |
| $\mathbf{I}$ | The secondary image |
| $\mathbf{I}_Y$ | The luminance component image |
| $k$ | The Number of ring-ribbons |
| $\mathbf{I}_{R(k)}$ | The ring-ribbons images |
| $\mathbf{I}_{Q(k)}$ | The quadtree images of ring-ribbons images |
| $L$ | The height and the width of image |
| $\tau$ | The first percentage of corner points |
| $\varsigma'$ | The screened corner points |
| $V_C$ | The variance conditional of quadtree partition |
| $\mathscr{F}_I, \mathscr{F}_{II}$ | Two feature fusion methods |
| $\xi$ | The threshold |
| $\eta$ | The length of sequence |

$$E_G(u,v) = \frac{E^{(1)}(u,v)}{\sum_{u=1}^{L}\sum_{v=1}^{L} E^{(1)}(u,v)}, \qquad (1)$$

$$E^{(1)}(u,v) = e^{\frac{-(u^2+v^2)}{2\sigma^2}}, \qquad (2)$$

where $E_G(u, v)$ denotes the values of the convolution mask at the coordinate $(u, v)$, $\sigma$ is a standard deviation of all elements in the mask. After the original image $\mathbf{I}_o$ is resized and smoothed, the secondary image $\mathbf{I}$ can be produced. In the following, the local and global feature extraction for texture and color information are conducted on the secondary image $\mathbf{I}$ to produce the image hash.

**2.2 Local Feature Extraction**

In this work, the secondary image $\mathbf{I}$ is divided into $N$ concentric ring-ribbons $\mathbf{R}_k$ with the equal area, $k = 1, 2, ..., N$, and then, the local textural feature are extracted based on QD, while the local color feature are extracted based on CVA.

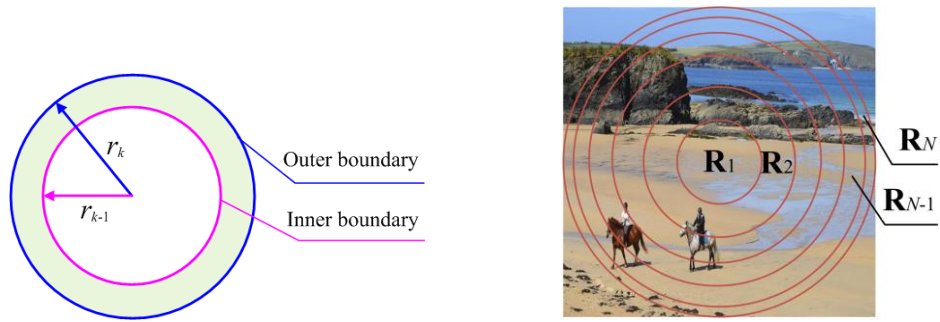

(a) Schematic diagram of one ring-ribbon    (b) Schematic diagram of $\mathbf{R}_k$

Fig. 3 An illustration of ring-ribbon.

*A. Ring-ribbon Division*

The ring-ribbon $\mathbf{R}_k$ is an annular object, which is encircled by the circumferences of two concentric rings, i.e., inner boundary and outer boundary, as illustrated in Fig. 3(a). Note that, the innermost ring-ribbon $\mathbf{R}_1$ is actually a circle. Denote the outer boundary of the ring-ribbon $\mathbf{R}_N$ as the inscribed circle of $\mathbf{I}$, as shown in Fig. 3(b). Thus, the radius $r_N$ of the outer boundary for $\mathbf{R}_N$ is:

$$r_N = \lfloor L/2 \rfloor, \tag{4}$$

where $\lfloor \cdot \rfloor$ means downward rounding. The area $M$ of the inscribed circle of $\mathbf{I}$ can be calculated as:

$$M = \pi r_N^2, \tag{5}$$

Hence, the area $\rho$ of each ring-ribbon $\mathbf{R}_k$ can be obtained as:

$$\rho = M / N. \tag{6}$$

Thus, for each ring-ribbon $\mathbf{R}_k$, the radius $r_k$ of outer boundary can be calculated by Eq. (10).

$$r_k = \begin{cases} \sqrt{\dfrac{\rho}{\pi}}, & \text{if } k = 1, \\ \sqrt{r_{k-1}^2 + \dfrac{\rho}{\pi}}, & \text{if } k > 1. \end{cases} \tag{10}$$

Let $(x_i, y_i)$ be the coordinate of the pixel $p_{i,j}$ in the secondary image $\mathbf{I}$ ($1 \leq i, j \leq L$). When $L$ is odd, the center coordinate $(x_o, y_o)$ of the image are considered as $x_o = (L + 1) / 2$, $y_o = (L + 1) / 2$. Otherwise, $x_o = (L / 2 + 0.5)$, $y_o = (L / 2 + 0.5)$. In order to form ring-ribbons with the equal area (consisting the same number of pixels) in the inscribed circle of the image $\mathbf{I}$, the distance $d_{i,j}$ from each pixel at $(x_i, y_i)$ to the image center $(x_o, y_o)$ are calculated:

$$d_{i,j} = \sqrt{(x_i - x_o)^2 + (y_j - y_o)^2}. \tag{11}$$

According to the relationship between the distances $d_{i,j}$ and the radius $r_k$, we can divide the image pixels into $N$ sets corresponding to $N$ ring-ribbons $\mathbf{R}_k$, $k = 1, 2, ..., N$, see Eq. (12).

$$\mathbf{R}_1 = \{p_{i,j} \mid d_{i,j} \leq r_1\}, \tag{12}$$

$$\mathbf{R}_k = \{p_{i,j} \mid r_{k-1} < d_{i,j} \leq r_k\}. \tag{13}$$

Obviously, when the image is rotated with different angles, the division of corresponding ring-ribbons remains unchanged, and each ring-ribbon still consists of the same pixels.

*B. Local Textural Feature Based on QD*

In the process of local textural feature extraction, the luminance component $\mathbf{R}_k^Y$ of $\mathbf{R}_k$, which can represent main visual contents of the ring-ribbon, is retrieved from YCbCr space of each $\mathbf{R}_k$ and then used for local textural feature extraction. The reasons for choosing YCbCr color space are analyzed in Section 4. The relationship of linear transformation from the RGB color space to the YCbCr color space is given in Eq. (3).

$$\begin{bmatrix} Y \\ C_b \\ C_r \end{bmatrix} = \begin{bmatrix} 0.2990 & 0.5870 & 0.1140 \\ -0.1687 & -0.3313 & 0.5000 \\ 0.5000 & -0.4187 & 0.8138 \end{bmatrix} \times \begin{bmatrix} R \\ G \\ B \end{bmatrix} + \begin{bmatrix} 0 \\ 128 \\ 128 \end{bmatrix}, \tag{3}$$

where $Y$ denotes the luminance, $C_b$ and $C_r$ denote color component differences for blue and red, respectively, and $R$, $G$, $B$ denote the red, green and blue components, respectively. Then, QD is conducted on each $\mathbf{R}_k^Y$ to obtain the corresponding QD result.

QD is an important image representation method, which recursively divides the two-dimensional space into different levels of tree structure. This recursion of division continues until the hierarchy of the tree reaches a certain depth or meets a certain requirement. As an example shown in Fig. 5, a typical quadtree structure is illustrated with a two-dimensional space structure of (a) and a storage structure of (b). Each quadtree node in (b) corresponds to a rectangular region in (a), which consists of the four rectangular regions of its four child nodes (if existing). There are $J + 1$ levels in the storage structure of (b), and Level 0 represents the root node of the quadtree, which corresponds to the rectangle with the longest side length in the space structure of (a). If the variance of the corresponding rectangle is greater than a pre-determined threshold $V_C$, the root node is first divided into four child nodes (denoted as Level 1), corresponding to four rectangles with the equal area in (a). Then, each child node is performed with the same QD process as described above. Finally, when all child nodes don't satisfy the decomposition condition, i.e., the variances of all child nodes are not greater than $V_C$, the QD process stops and the whole quadtree is generated.

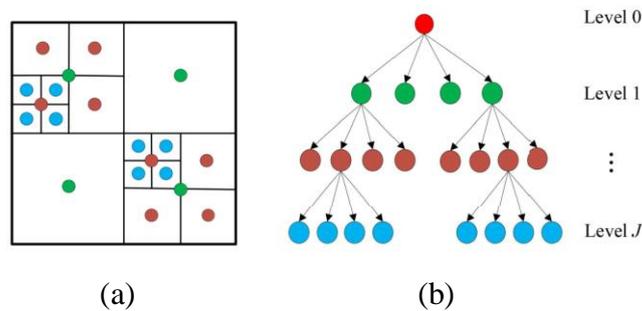

(a)          (b)

Fig. 5 An illustration of Quadtree structure. (a) Two-dimensional space structure, (b) Storage structure.

Our hashing scheme extracts the local textural features from the divided ring-ribbons $\mathbf{R}_k^Y$ ($k = 1, 2, ..., N$). Before textural feature extraction for each $\mathbf{R}_k^Y$, we set the image region, expect the current ring-ribbon $\mathbf{R}_k^Y$, as white, and then conduct QD for $\mathbf{R}_k^Y$ within the image sized $L \times L$. In our scheme, the number of decomposition times for each ring-ribbon image is indicated as the local textural feature. For example, the number of decomposition times for Fig. 5(a) is 5, that is to say, five rectangles in (a) and the corresponding five nodes in (b) are decomposed.

Fig. 6 shows an example of QD results with different values of $V_C$ and $N$ for the ring-ribbons $\mathbf{R}_N$ of the image in (a). Fig. 6(b) are $\mathbf{R}_N$ ($N = 3$) divided from $\mathbf{I}$, Figs. 6(c)-(e) are the QD results with $V_C = 10, 30, 50$, respectively. As shown in Fig. 6(c), when $V_C$ is small, some acceptable slight differences of textures are decomposed. On the other hand, when $V_C$ is large, the decomposition of some nodes with different textures will be ignored, which can be found in Fig. 6(e). In addition to different values of $V_C$ can present different QD results, the $N$ of RN will also affect the classification performance of the scheme. Figs. 6(f) and (g) are the QD results with $N = 5$ and $N = 10$, respectively, whose values of $V_C$ are all 30. It can be observed that the greater the $N$ of $\mathbf{R}_N$, the fewer number of pixels and the information contained in the $\mathbf{R}_N$. As shown in Fig. 6(g), when $N = 10$, part of the QD results are lost. Clearly, we will comprehensively consider different values of $V_C$ and $N$ to get the well performance of proposed scheme.

The analysis of the parameter $V_C$ and $N$ are discussed detailedly in Section IV. Thus, we can acquire one feature vector of local texture, i.e., $\mathbf{H}_Q$, with $N$ elements corresponding to the $N$ ring-ribbons, i.e., $\mathbf{R}_1, \mathbf{R}_2, \ldots, \mathbf{R}_N$, see Eq. (14).

$$\mathbf{H}_Q = [H_Q^{(1)}, H_Q^{(2)}, \ldots, H_Q^{(N)}], \tag{14}$$

where $H_Q^{(i)}$ is the value of decomposition times for the $k$th ring-ribbon $\mathbf{R}_k^Y$.

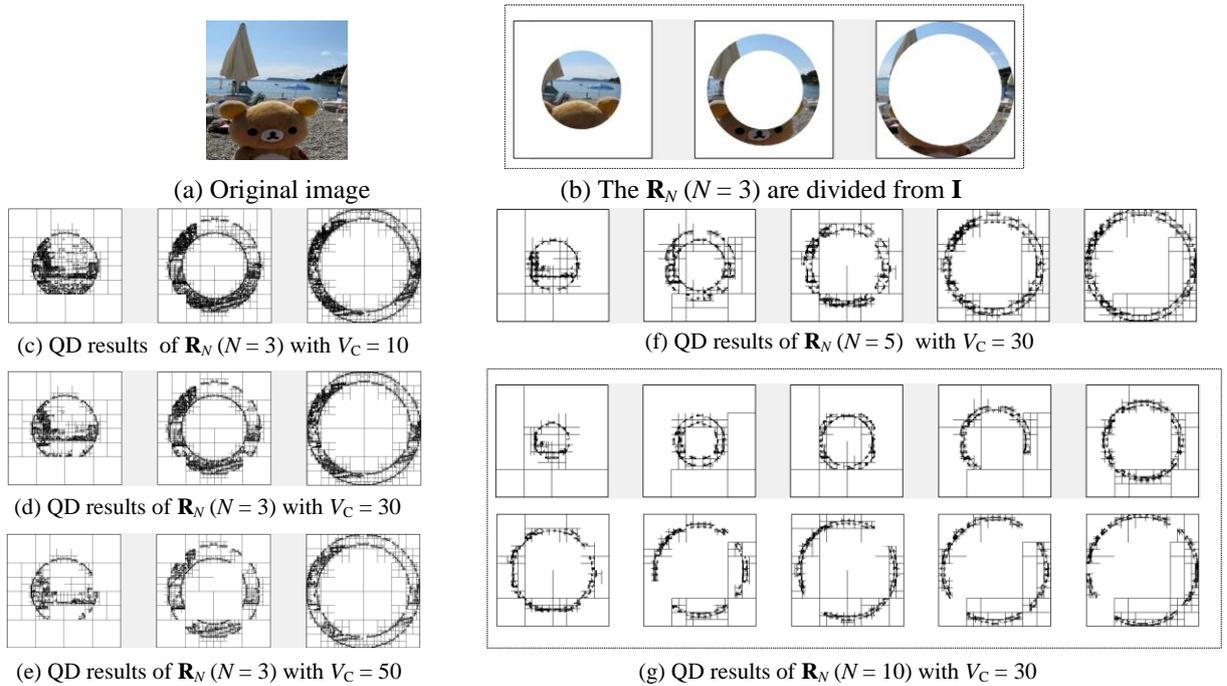

(a) Original image  
(b) The $\mathbf{R}_N$ ($N = 3$) are divided from $\mathbf{I}$  
(c) QD results of $\mathbf{R}_N$ ($N = 3$) with $V_C = 10$  
(f) QD results of $\mathbf{R}_N$ ($N = 5$) with $V_C = 30$  
(d) QD results of $\mathbf{R}_N$ ($N = 3$) with $V_C = 30$  
(e) QD results of $\mathbf{R}_N$ ($N = 3$) with $V_C = 50$  
(g) QD results of $\mathbf{R}_N$ ($N = 10$) with $V_C = 30$

Fig. 6 The QD results with different values of $V_C$ and $N$.

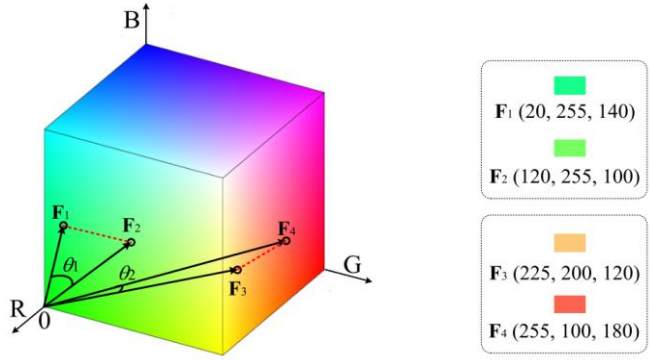

Fig. 7 CVA comparison of two color pairs with the same $D_E$.

*C. Local Color Feature Based on CVA*

In addition to local textural features based on the ring-ribbon quadtree decomposition, local color features are also considered in our scheme. Denote $\mathbf{F}_1 = [F_r^{(1)}, F_g^{(1)}, F_b^{(1)}]$ and $\mathbf{F}_2 = [F_r^{(2)}, F_g^{(2)}, F_b^{(2)}]$ as two different colors in RGB space, where $F_r^{(v)}$, $F_g^{(v)}$ and $F_b^{(v)}$ ($v = 1, 2$) are the red, green and blue components of the pixel. The Euclidean distance $D_E$ of $\mathbf{F}_1$ and $\mathbf{F}_2$ can be easily calculated as:

$$D_E(\mathbf{F}_1, \mathbf{F}_2) = [(F_r^{(1)} - F_r^{(2)})^2 + (F_g^{(1)} - F_g^{(2)})^2 + (F_b^{(1)} - F_b^{(2)})^2]^{\frac{1}{2}}, \quad (15)$$

However, as shown in Fig. 7, the similarity of the color pair ($\mathbf{F}_1$, $\mathbf{F}_2$) is obviously higher than ($\mathbf{F}_3$, $\mathbf{F}_4$) in human visual perception, but the two Euclidean distances $D_E$ for ($\mathbf{F}_1$, $\mathbf{F}_2$) and ($\mathbf{F}_3$, $\mathbf{F}_4$) are the same, which means Euclidean distance cannot measure the color difference accurately. Therefore, in our scheme, the Euclidean distance is replaced by the CVA to measure color difference and extract color feature. CVA of $\mathbf{F}_1$ and $\mathbf{F}_2$ can be calculated by see Eq. (16).

$$\theta = \arcsin\left(1 - \frac{(\mathbf{F}_1^T \mathbf{F}_2)^2}{\mathbf{F}_1^T \mathbf{F}_1 \mathbf{F}_2^T \mathbf{F}_2}\right)^{\frac{1}{2}}. \quad (16)$$

In order to control the value range into [0, 1], we utilize the sine value of CVA instead of the angle $\theta$ in our scheme, see Eq. (17).

$$\sin\theta = \left(1 - \frac{(\mathbf{F}_1^T \mathbf{F}_2)^2}{\mathbf{F}_1^T \mathbf{F}_1 \mathbf{F}_2^T \mathbf{F}_2}\right)^{\frac{1}{2}}. \quad (17)$$

In order to extract local color features, Harris corner points on the outer boundary of each ring-ribbon $\mathbf{R}_k$ are first detected. Denote the number of the detected Harris corner points on the outer boundary of each ring-ribbon $\mathbf{R}_k$ as $\varsigma_b^{(k)}$, $k = 1, 2, \ldots, N$, and sort the $\varsigma_b^{(k)}$ corner points in the descending order of intensity for each $\mathbf{R}_k$. To achieve satisfactory robustness, the first $\tau$ percentage of the $\varsigma_b^{(k)}$ corner points are selected. Thus, the number $\varsigma^{(k)'}$ of corner points used to extract local color feature for each $\mathbf{R}_k$ is:

$$\varsigma^{(k)'} = \varsigma_b^{(k)} \times \tau. \tag{22}$$

Figure 8 shows an example of the selected Harris corner points of three ring-ribbons in an image. To make the display clearer, as shown in Fig. 8(a), all Harris corner points on the outer boundary of each ring-ribbon $\mathbf{R}_k$ are marked as red dots. The red dots in Fig. 8(b) represents the first $\tau$ ($\tau = 35$) percentage of corner points.

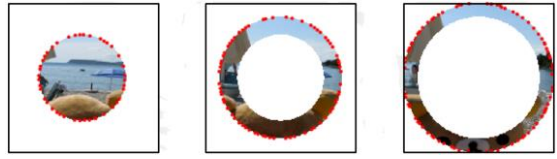

(a) All Harris corner points on the outer boundary of each $\mathbf{R}_k$

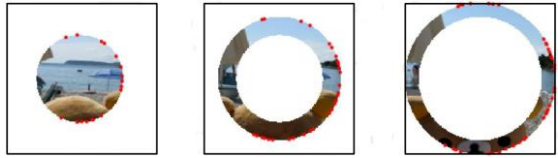

(b) The first $\tau$ ($\tau = 35$) percentage of $\mathbf{R}_k$ corner points

Fig. 8 Test images and corner points of outer boundaries.

Then, we calculate the CVA for these selected Harris corner points on each ring-ribbon $\mathbf{R}_k$. Denote $\mathbf{P}_{ref} = [R_{ref}, G_{ref}, B_{ref}]$ as the reference color in our CVA calculation, where $R_{ref}, G_{ref}, B_{ref}$ are the average values for the Red channel, Green channel and Blue channel of the secondary image $\mathbf{I}$, respectively. For each ring-ribbon $\mathbf{R}_k$, the $\varsigma^{(k)'}$ selected Harris corner points are denoted as $\mathbf{p}_{(1)}, \mathbf{p}_{(2)}, \ldots, \mathbf{p}_{(\varsigma^{(k)'})}$, and the CVA of each point $\mathbf{p}_{(i)}$ can be computed as:

$$C_i = \left(1 - \frac{(\mathbf{p}_{(i)}^T \mathbf{P}_{ref})^2}{\mathbf{p}_{(i)}^T \mathbf{p}_{(i)} \mathbf{P}_{ref}^T \mathbf{P}_{ref}}\right)^{\frac{1}{2}}. \tag{23}$$

After the CVA values of all selected Harris corner points are calculated, for each ring-ribbon $\mathbf{R}_k$, the variance $V_{(k)}^2$ of the $\varsigma^{(k)'}$ CVA values can be obtained:

$$V_{(k)}^2 = \frac{1}{\varsigma^{(k)'}} \cdot \sum_{i=1}^{\varsigma^{(k)'}} [(\frac{1}{\varsigma^{(k)'}} \cdot \sum_{i=1}^{\varsigma^{(k)'}} C_i) - C_i]^2, \tag{24}$$

Thus, all $N$ CVA variances corresponding to $N$ ring-ribbons can be acquired and used as the local color feature vector $\mathbf{H}_C$, see Eq. (25).

$$\mathbf{H}_C = [H_C^{(1)}, H_C^{(2)}, \ldots, H_C^{(N)}]. \tag{25}$$

## 2.4 Global Feature Extraction

Beside local features based on a series of ring-ribbons, we also consider the global features of the secondary image **I**. In our scheme, the gray-level co-occurrence matrix (GLCM) and the color low-order moments (CLMs) are utilized to describe the global features for texture and structure, respectively.

As for global texture feature, the GLCM can reflect the comprehensive information of the image **I** grayscale about the direction and the adjacent interval. A set of "point pairs" is composed of any point $(x, y)$ and adjacent points $(x + a, y + b)$, where $a$ and $b$ define the position directions $\theta$ and distance $d$, specifically expressed as

$$\begin{cases} \{(a=0, |b|=d)\}, & \theta = 0° \\ \{(a=d, b=-d), (a=-d, b=d)\}, & \theta = 45° \\ \{(a=d, |b|=0)\}, & \theta = 90° \\ \{(a=d, b=-d), (a=-d, b=d)\}, & \theta = 135° \end{cases}, \quad (26)$$

where $\theta = 0°$ and $\theta = 90°$ represent the adjacent horizontal and vertical directions, adjacent in the diagonal directions are represented by 45° and 135°. Suppose the gray value of the "point pair" is $(g_1, g_2)$, assuming the maximum gray level of the image is $g_{max}$, then there are $g_{max} \times g_{max}$ combinations of $g_1$ and $g_2$. Furthermore, the global texture information can be represented by a relative frequency matrix (FM), which is the frequency of "point pairs" with distance $d$. GLCM is first calculated for the luminance component of the secondary image **I**, see Eqs. (27)-(28).

$$G_{g_1, g_2}(d, \theta) = \frac{\#C_{g_1, g_2}(d, \theta)}{\#S}, \quad (27)$$

$$\#S = \begin{cases} 2L_m(L_n - d), & \theta = 0° \\ 2L_n(L_m - 1), & \theta = 90° \\ 2(L_m - d)(L_n - d), & \theta = 45° \text{ or } \theta = 135° \end{cases}, \quad (28)$$

where $G_{g_1, g_2}(d, \theta)$ is the probability of simultaneous occurrence of "point pairs $(g_1, g_2)$" in the $\theta$ direction, $\#C_{g_1, g_2}(d, \theta)$ is the number of times "point pairs $(g_1, g_2)$" appear simultaneously in $\theta$ direction (# represents number), $L_m$ is the number of rows of the original image, $L_n$ is the number of columns of the original image, and $\#S$ is the total number of times of all "point pairs" with distance $d$ in the $\theta$ direction. Suppose that the given statistical direction $\theta = 0°$, the interval distance $d = 1$, and $g_{max} = 6$. The image (size of 6 × 6) is converted into its corresponding "point pair" frequency matrix FM, an example is shown in Fig. 9. Fig. 9 (a) is the gray level matrix of the test image, and Fig.9 (b) is the frequency of gray level "point pairs" in $\theta = 0°$. In terms of "point pair (1, 1)" in (a), the corresponding value in (b) is 4, which means that only 4 pairs of pixels with a gray level of 1 are horizontally adjacent, and $G_{1,1}(1, 0°) = 0.0667$.

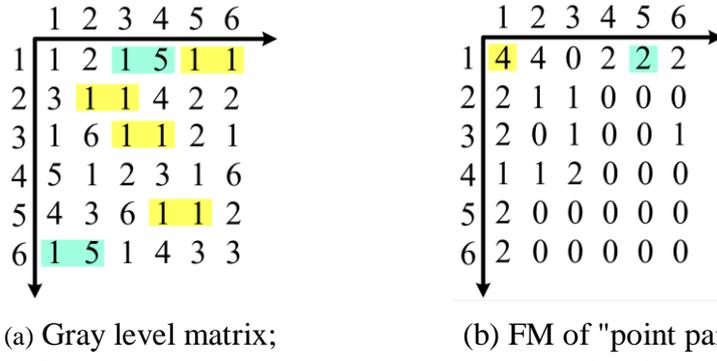

(a) Gray level matrix;   (b) FM of "point pairs".

Fig. 9 (a) The gray level matrix of the test image, (b) FM of gray level "point pairs" in the $\theta = 0°$ direction.

And three scalars of GLCM, i.e., correlation, contrast and entropy, are characterized as follows:

$$Cor = \frac{\sum_i \sum_j (i-\bar{x})(j-\bar{y})G(i,j)}{\sigma_x \sigma_y}, \quad s.t. \begin{cases} \bar{x} = \sum_i i \sum_j G(i,j) \\ \bar{y} = \sum_i j \sum_j G(i,j) \\ \sigma_x^2 = \sum_i (i-\bar{x})^2 \sum_j G(i,j) \\ \sigma_y^2 = \sum_j (j-\bar{x})^2 \sum_j G(i,j) \end{cases} \quad (24)$$

where $G(i, j)$ is the GLCM element value in the $i$-th row and $j$-th column, and the value $Cor$ of correlation measures the similarity degree of spatial GLCM elements in horizontal and vertical directions.

$$Con = \sum_i \sum_j (i-j)^2 G(i,j), \quad (25)$$

where the value $Con$ of contrast reflects the sharpness of the GLCM that is related with the texture depth of the image.

$$Ent = -\sum_i \sum_j G(i,j) \lg G(i,j), \quad (26)$$

where the value $Ent$ of entropy is a measure of image information that represents the degree of non-uniformity and complexity of image textures. When all elements in the GLCM have greater randomness, the value of $Ent$ is larger. Thus, these three calculated scalars are concatenated to produce $\mathbf{Z}_G = [Cor, Con, Ent]$ as the global texture feature.

As for global color feature, since color distribution information is mainly related with the color low-order moments (CLMs), hence, we utilize the first-order moment $C_\Theta$, the second-order moment $C_\Psi$ and the third-order moment $C_\Xi$ to describe global color distribution of the image, see Eqs. (27)-(29).

$$C_\Theta = \frac{1}{N} \sum_{j=1}^{N} A_{ij}, \quad (27)$$

$$C_\Psi = \left[ \frac{1}{N} \sum_{j=1}^{N} (A_{ij} - C_\Theta)^2 \right]^{\frac{1}{2}}, \quad (28)$$

$$C_\Xi = \left[ \frac{1}{N} \sum_{j=1}^{N} (A_{ij} - C_\Theta)^3 \right]^{\frac{1}{3}}, \tag{29}$$

where $C_\Theta$ defines the average intensity of the image, $C_\Psi$ reflects the color variance, and $C_\Xi$ defines the skewness of the image. Thus, the global color feature can be represented by $\mathbf{C}_G = [C_\Theta, C_\Psi, C_\Xi]$.

## 2.5 Hash Generation

After the stage of feature extraction, the local feature vectors for texture and color, i.e., $\mathbf{H}_Q$ based on QD and $\mathbf{H}_C$ based on CVA, are obtained, and the global feature vectors for texture and color, i.e., $\mathbf{Z}_G$ based on GLCM and $\mathbf{C}_G$ based on CLMs, are also acquired. Then, these extracted feature vectors are used to generate the final image hash. Two hash generation methods are given and discussed as follows.

*A. Hash Generation with Direct Concatenation*

The textural feature vectors ($\mathbf{H}_Q$ and $\mathbf{Z}_G$) and color feature vectors ($\mathbf{H}_C$ and $\mathbf{C}_G$) are directly concatenated as:

$$\mathbf{H}_{int} = [\mathbf{H}_Q, \mathbf{Z}_G, \mathbf{H}_C, \mathbf{C}_G], \tag{30}$$

The lengths of the vectors $\mathbf{H}_Q$, $\mathbf{Z}_G$, $\mathbf{H}_C$ and $\mathbf{C}_G$ are $N$, $N$, 3 and 3, respectively, where $N$ is the number of the divided ring-ribbons. Then, according to the secret key $S_{key}$, all elements in $\mathbf{H}_{int}$ is scrambled to produce the final hash $\mathbf{H}$, and the length of final hash $\mathbf{H}$ is $(2N + 6)$.

*B. Hash Generation with Canonical Correlation Analysis*

As described in part A of Section 2.5, two or more feature vectors are concatenated in a certain order to obtain final hashing. The operation of direct concatenation has the following disadvantages: the concatenated hash sequence ignores the internal relationship between each feature vector and feature redundancy is retained. Here, we utilize CCA fuse feature vectors. CCA is essentially a multivariate statistical analysis method that can use the relationship between comprehensive variable pairs to reflect the overall correlation of the two sets of indicators.

For the same image, there is a certain complementarity and redundancy between the features extracted locally and globally. Therefore, we fuse texture feature and color feature by CCA, and then the fused, correlated features are used for image classification. Main steps are as follows: (1) Two sets of textural features based on local QD and global GLCM are denoted as $\mathbf{H}_Q$ and $\mathbf{Z}_G$; (2) The other two sets of color features based on local CVA and global CLMs are denoted as $\mathbf{H}_C$ and $\mathbf{C}_G$; (3) Take $\mathbf{H}_1 = [\mathbf{H}_Q, \mathbf{Z}_G]$, $\mathbf{H}_2 = [\mathbf{H}_C, \mathbf{C}_G]$ as the inputs of CCA, find two sets of projection matrixes **a**, **b**; (4) The two sets of related features obtained are fused. Details are introduced in the following.

In linear regression, a straight line is used to fit sample points. Let two feature vectors $\mathbf{H}_1 = (h_1^{(1)}, h_2^{(1)}, \ldots, h_N^{(1)})^T$, $\mathbf{H}_2 = (h_1^{(2)}, h_2^{(2)}, \ldots, h_N^{(2)})^T$, where $N$ represent the dimensions of two feature vectors. The $\mathbf{H}_1$ and $\mathbf{H}_2$ can establish the equation $\mathbf{H}_2 = \mathbf{W}\mathbf{H}_1$ to find the linear relationship, see Eq. (31).

$$\begin{bmatrix} h_1^{(2)} \\ h_2^{(2)} \\ \vdots \\ h_N^{(2)} \end{bmatrix} = \begin{bmatrix} w_{11} & w_{12} & \ldots & w_{1N} \\ w_{21} & w_{22} & \ldots & w_{2N} \\ \vdots & \vdots & \vdots & \vdots \\ w_{N1} & w_{N2} & \ldots & w_{NN} \end{bmatrix} \begin{bmatrix} h_1^{(1)} \\ h_2^{(1)} \\ \vdots \\ h_N^{(1)} \end{bmatrix}, \quad (31)$$

where $y_i = w_i^T \mathbf{H}_1$, clearly, each element of $\mathbf{H}_2$ is associated with all the characteristics of $\mathbf{H}_1$, but the correlation between $\mathbf{H}_1$ and $\mathbf{H}_2$ internal variables are not considered. Let's express this relation in another way, $\mathbf{H}_1$ and $\mathbf{H}_2$ are both regarded as a whole, and consider the relationship between these two wholes. Defines $<\cdot, \cdot>$ as inner product of the vectors, $<a_1, \mathbf{H}_1>$ represents the projection of $\mathbf{H}_1$ in the $a_1$ direction, value of **v** is obtained by $<b_1, \mathbf{H}_2>$. Let the $(h_1^{(1)}, h_2^{(1)}, \ldots, h_N^{(1)})^T$ and $(h_1^{(2)}, h_2^{(2)}, \ldots, h_N^{(2)})^T$ are expressed as follows.

$$\mathbf{u} = (<a_1, h_1^{(1)}>, <a_2, h_2^{(1)}>, \ldots, <a_N, h_N^{(1)}>), \quad (32)$$

$$\mathbf{v} = (<b_1, h_1^{(2)}>, <b_2, h_2^{(2)}>, \ldots, <b_N, h_N^{(2)}>). \quad (33)$$

In order to measure the correlation between **u** and **v**, we expect to find the optimal solution $\mathbf{a} = (a_1, a_2, \ldots, a_N)^T$, $\mathbf{b} = (b_1, b_2, \ldots, b_N)^T$, which can make correlation $Corr(\mathbf{u}, \mathbf{v})$ get the maximum. The obtained **a** and **b** are the projection matrixes that make **u** and **v** have the greatest correlation, see Eqs. (34)-(37).

$$Var(\mathbf{u}) = \mathbf{a}^T cov(\mathbf{u})\mathbf{a} = \mathbf{a}^T S_{\mathbf{H}_1\mathbf{H}_1} \mathbf{a}, \quad (34)$$

$$Var(\mathbf{v}) = \mathbf{b}^T cov(\mathbf{v})\mathbf{b} = \mathbf{b}^T S_{\mathbf{H}_2\mathbf{H}_2} \mathbf{b}, \quad (35)$$

$$cov(\mathbf{u}, \mathbf{v}) = \mathbf{a}^T cov(\mathbf{uv})\mathbf{b} = \mathbf{a}^T S_{\mathbf{H}_1\mathbf{H}_2} \mathbf{b}, \quad (36)$$

$$\rho = \max Corr(\mathbf{u}, \mathbf{v}) = \frac{cov(\mathbf{u}, \mathbf{v})}{\sqrt{Var(\mathbf{u})}\sqrt{Var(\mathbf{v})}} = \frac{\mathbf{a}^T S_{\mathbf{H}_1\mathbf{H}_2} \mathbf{b}}{\sqrt{\mathbf{a}^T S_{\mathbf{H}_1\mathbf{H}_1} \mathbf{a}} \sqrt{\mathbf{b}^T S_{\mathbf{H}_2\mathbf{H}_2} \mathbf{b}}}, \quad (37)$$

where $S_{\mathbf{H}_1\mathbf{H}_1}$, $S_{\mathbf{H}_2\mathbf{H}_2}$ represent the two variance matrixes corresponding to $\mathbf{H}_1$ and $\mathbf{H}_2$, $S_{\mathbf{H}_1\mathbf{H}_2}$ is the covariance matrix of $\mathbf{H}_1$ and $\mathbf{H}_2$, $S_{\mathbf{H}_2\mathbf{H}_1}$ is equal to $S_{\mathbf{H}_1\mathbf{H}_2}$. In other words, the goal of our CCA scheme is finally transformed into a convex optimization process. In addition, **a**, **b** are the projection matrixes or linear coefficients in the dimensionality reduction process. The constraints of this optimization problem are describe as follows.

$$\arg\max_{\mathbf{a}_1,\mathbf{b}_1}\left\{\mathbf{a}^T S_{\mathbf{H}_1\mathbf{H}_2}\mathbf{b}\right\},$$

$$\text{s.t.}\begin{cases}\mathbf{a}^T S_{\mathbf{H}_1\mathbf{H}_1}\mathbf{a}=1\\ \mathbf{b}^T S_{\mathbf{H}_2\mathbf{H}_2}\mathbf{b}=1\end{cases}. \tag{38}$$

Maximize the numerator of formula Eq. (37). The Lagrangian equation is introduced to calculate conditional extreme value, see Eq. (39).

$$\Phi(\mathbf{a},\mathbf{b})=\mathbf{a}^T S_{\mathbf{H}_1\mathbf{H}_2}\mathbf{b}-\frac{\lambda_a}{2}(\mathbf{a}^T S_{\mathbf{H}_1\mathbf{H}_1}\mathbf{a}-1)-\frac{\lambda_b}{2}(\mathbf{b}^T S_{\mathbf{H}_2\mathbf{H}_2}\mathbf{b}-1), \tag{39}$$

where $\lambda_a$ and $\lambda_b$ are Lagrangian multipliers. Calculate the partial derivatives of Eq. (39) with **a** and **b**, after setting the derivative to 0, the system of equations is obtained as follows.

$$\frac{\partial \Phi}{\partial \mathbf{a}}=S_{\mathbf{H}_1\mathbf{H}_2}\mathbf{b}-\lambda_a S_{\mathbf{H}_1\mathbf{H}_1}\mathbf{a}=0, \tag{40}$$

$$\frac{\partial \Phi}{\partial \mathbf{b}}=S_{\mathbf{H}_2\mathbf{H}_1}\mathbf{a}-\lambda_b S_{\mathbf{H}_2\mathbf{H}_2}\mathbf{b}=0, \tag{41}$$

Multiply Eq. (40) left by $\mathbf{a}^T$, Eq. (41) left by $\mathbf{b}^T$, according to constraints, we can get Eq. (42).

$$\begin{aligned}0&=\mathbf{a}^T S_{\mathbf{H}_1\mathbf{H}_2}\mathbf{b}-\mathbf{a}^T\lambda_a S_{\mathbf{H}_1\mathbf{H}_1}\mathbf{a}-\mathbf{b}^T S_{\mathbf{H}_2\mathbf{H}_1}\mathbf{a}+\mathbf{b}^T\lambda_b S_{\mathbf{H}_2\mathbf{H}_2}\mathbf{b}\\ &=\lambda_b\mathbf{b}^T S_{\mathbf{H}_2\mathbf{H}_2}\mathbf{b}-\lambda_a\mathbf{a}^T S_{\mathbf{H}_1\mathbf{H}_1}\mathbf{a},\end{aligned} \tag{42}$$

together with the constraints, that

$$\lambda=\lambda_a=\lambda_b=\mathbf{b}^T S_{\mathbf{H}_2\mathbf{H}_1}\mathbf{a}. \tag{43}$$

That is means, the calculated $\lambda$ is *Corr* (**u**, **v**), and then, assuming $S_{H_2H_2}$ in Eq. (41) is invertible,

$$\mathbf{b}=\frac{1}{\lambda}S^{-1}_{\mathbf{h}_2\mathbf{h}_2}S_{\mathbf{H}_2\mathbf{H}_1}\mathbf{a}, \tag{44}$$

and Eq. (45) is substituting in Eq. (40) and Eq. (41), we have

$$S_{\mathbf{H}_1\mathbf{H}_2}S^{-1}_{\mathbf{H}_2\mathbf{H}_2}S_{\mathbf{H}_2\mathbf{H}_1}\mathbf{a}=\lambda^2 S_{\mathbf{H}_1\mathbf{H}_1}\mathbf{a}. \tag{45}$$

$$S_{\mathbf{H}_2\mathbf{H}_1}S^{-1}_{\mathbf{H}_1\mathbf{H}_1}S_{\mathbf{H}_1\mathbf{H}_2}\mathbf{b}=\lambda^2 S_{\mathbf{H}_2\mathbf{H}_2}\mathbf{b}. \tag{46}$$

Continue to simplify Eqs. (45)-(46) and write them in matrix form to get as follows:

$$\begin{bmatrix}S^{-1}_{\mathbf{H}_1\mathbf{H}_1}S_{\mathbf{H}_1\mathbf{H}_2} & 0\\ 0 & S^{-1}_{\mathbf{H}_2\mathbf{H}_2}S_{\mathbf{H}_2\mathbf{H}_1}\end{bmatrix}\begin{bmatrix}S^{-1}_{\mathbf{H}_2\mathbf{H}_2}S_{\mathbf{H}_2\mathbf{H}_1} & 0\\ 0 & S^{-1}_{\mathbf{H}_1\mathbf{H}_1}S_{\mathbf{H}_1\mathbf{H}_2}\end{bmatrix}\begin{bmatrix}\mathbf{a}\\ \mathbf{b}\end{bmatrix}=\lambda^2\begin{bmatrix}\mathbf{a}\\ \mathbf{b}\end{bmatrix}. \tag{47}$$

The conditional extreme is converted to $\lambda^2$ and eigenvectors **a**, **b**. The second canonical variable of demand solution reflects the correlation between the two sets of multivariate. The calculation steps are the same as those in the first group, except that $\lambda$' takes the second largest eigenvalue. Solving for more typical variables reflects the correlation between two sets of multiple variables. Take the eigenvector corresponding to the largest eigenvalue of the previous $e$ to form the projection axis, $\mathbf{a} = (a_1, a_2, \ldots, a_e)$ and $\mathbf{b} = (b_1, b_2, \ldots, b_e)$ can be found. The method is used for features fusion as follows:

$$\mathbf{H}_{1pro} = \mathbf{a}^T \mathbf{H}_1, \tag{48}$$

$$\mathbf{H}_{2pro} = \mathbf{b}^T \mathbf{H}_2, \tag{49}$$

$$\mathbf{F} = \mathbf{H}_{1pro} + \mathbf{H}_{2pro} = \mathbf{a}^T \mathbf{H}_1 + \mathbf{b}^T \mathbf{H}_2. \tag{50}$$

The CCA feature fusion strategy is applied to the image classification task, which can eliminate redundant information while maximizing the correlation between different feature vectors.

*C. Performance comparison of two hash generation methods*

In this section, the performance comparison of two hash generation methods is introduced. For fair comparison, the length of the CCA fusion feature is consistent with the direct concatenation hash. The detailed parameter analysis for feature fusion with CCA are discussed in Section 4.

Here, Receiver operating characteristics (ROC) curve is utilized as the decision tool of performance analysis and comparison for the two hash generation methods, i.e., direct concatenation and CCA fusion. ROC curve is a very important and common statistical analysis method, which is used to objectively judge the classification and the quality of the test results. In the ROC curve, the corresponding false positive rate ($P_{\text{FPR}}$) value of each point is the abscissa and true positive rate ($P_{\text{TPR}}$) value is the ordinate. Two of the most important formulas are as follows.

$$P_{\text{TPR}} = \frac{\omega_1}{\varepsilon_1}, \tag{51}$$

$$P_{\text{FPR}} = \frac{\omega_2}{\varepsilon_2}, \tag{52}$$

where $\varepsilon_1$ is the number of all similar image pairs, $\varepsilon_2$ is the number of all different image pairs, $\omega_1$ is the number of similar images correctly judged as similar, and $\omega_2$ is the number of different images judged as similar images. In the ROC curve, the $P_{\text{FPR}}$ value corresponding to each point is the abscissa, and the $P_{\text{TPR}}$ value is the ordinate. The above two indicators are used in image classification. The important task is to judge similar pictures, that is, the curve has preferable classification performance where is nearer to the top-left corner.

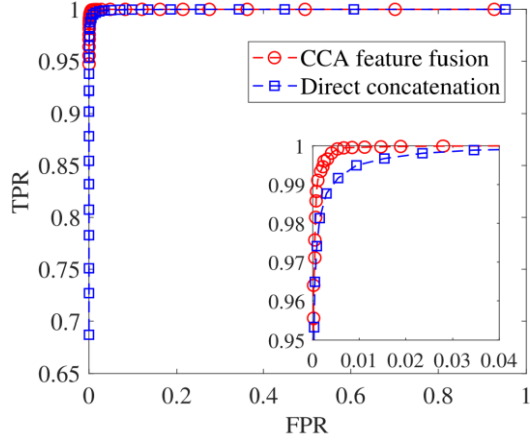

Fig. 10 ROC curves of traditional serial connection and CCA feature fusion.

We randomly collected 1,000 color images from the widely used COCO image database [27] for performance analysis. Those images include various categories, such as human, animals, buildings, natural scenery. The process of generating classification test images is described in section 3. And the parameters used in the experiment are listed in Table 2. The ROC curves of direct concatenation and CCA feature fusion is shown in Fig. 10. The extracted features can both obtain excellent image classification performance according to the direct concatenation and CCA feature fusion method, but the classification effect of CCA feature fusion is better. The results show that the CCA feature fusion method can be used as a simple and effective method for feature fusion in image re-classification tasks.

## 3. Experimental Results

A large number of experiments were conducted to verify the capabilities of our scheme in terms of robustness, discrimination and security.

Table 2 The parameter settings of our scheme

| Main stages | Parameter settings | Direct concatenation | CCA fusion |
|---|---|---|---|
| Pre-processing | Image normalized size $L \times L$ | $256 \times 256$ | $512 \times 512$ |
| Feature Extraction | Ring-ribbon number $k$ | 32 | 67 |
| | First $\tau$ percentage of corner points number | 0.4 | 0.4 |
| | Variance condition $V_C$ | 40 | 14 |
| Hash Generation | Length $N$ of $\mathbf{H}_Q$ | 32 | 67 |
| | Length $N_2$ of $\mathbf{H}_C$ | 32 | 67 |
| | Length of $\mathbf{Z}_G$ | 3 | 3 |
| | Length of $\mathbf{C}_G$ | 3 | 3 |
| | Length of $\mathbf{H}$ | 70 | 70 |

## 3.1 Measure Similarity

In our scheme, the elements of feature vectors are all decimal numbers. Therefore, both correlation coefficient and Euclidean distance can be utilized to measure the similarity of hash sequences.

*A. Correlation Coefficient*

Correlation coefficient is widely used to measure the degree of linear correlation between two hash sequences $\alpha'$ and $\alpha''$, its main idea is to find the difference between two sequences, which can also be said to be a measure of similarity. For the calculation of the similarity between two equal-length code strings, see Eq. (53).

$$\Omega(\alpha', \alpha'') = \frac{\sum_{i=1}^{T}(\alpha_i' - \overline{\alpha'}) \cdot (\alpha_i'' - \overline{\alpha''})}{\sqrt{\sum_{i=1}^{T}(\alpha_i' - \overline{\alpha'}) \cdot \sum_{i=1}^{T}(\alpha_i'' - \overline{\alpha''})} + \Delta o}, \quad (53)$$

where $T$ is the length of sequence, $\overline{\alpha'}$ and $\overline{\alpha''}$ are the mean values of $\alpha'$ and $\alpha''$, $\Delta o$ is an extremely small constant, which prevents the denominator part from being zero.

*B. Euclidean Distance*

Euclidean distance is also called $L_2$ norm, which measures the absolute distance between two points in *n*-dimensional space or the natural length of a vector (the distance from the point to the origin). The Euclidean distance was used to measure the similarity between two sequences $\beta'$ and $\beta''$, see Eq. (54).

$$\mho(\beta', \beta'') = \sqrt{\sum_{i=1}^{T}(h_\beta^{(i)}{'} - h_\beta^{(i)}{''})^2}, \quad (54)$$

where $\beta'$ and $\beta''$ are the hashes generated by the two images, $h_\beta^{(i)}{'}$ and $h_\beta^{(i)}{''}$ are the corresponding values of hashes, $T$ is the length of sequence. When the value of Euclidean distance exceeds the pre-determined threshold $\xi$, the two images are considered to be different or have been tampered.

*C. Similarity of Hash Sequence*

The similarity measurement of hash sequences is the basis of image classification. There is no optimal or general measurement criterion, it is necessary to choose a suitable method according to the specific problem. The local feature vectors of our scheme are composed of the number of each ring-ribbon decomposition times and the variances of CVA. The values of CVA variance are smaller, all within the range of [0, 1]. But the values of QD times are significantly greater than them. Therefore, the trend of the hash sequences of all images are bound to be the same.

Observed from Fig. 11, the Euclidean distance is more suitable for measuring the hash similarity of the proposed method. Since all the eigenvalues obtained are decimal numbers and are not converted into binary numbers, the hash similarity is more scientific.

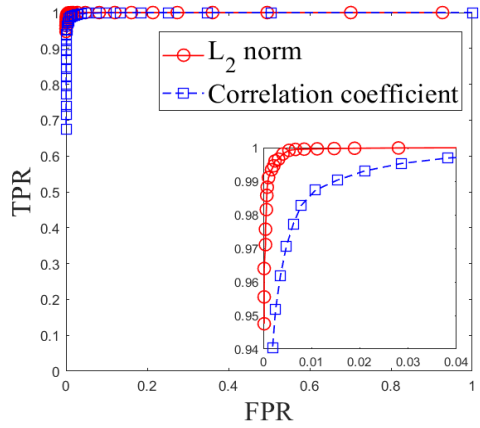

Figure 11 ROC curves of similarity measurement of hash sequence.

**3.2 Perceptual Robustness**

In this section, in order to verify the authenticity and reliability of the experimental results, robustness attacks were carried out by eight common content-preserving manipulations. StirMark [28] and MATLAB are selected as tools for robustness attacks in our scheme. The manipulations include scaling, JPEG compression, circular blurring, small angles rotation and cropping (keep the size of $512 \times 512$), and large angles rotation, etc. The large angles rotation will significantly increase the sizes of the original images, and the surrounding regions of these images are filled with black or white pixels. Therefore, when measuring the difference between pairs of large angles rotation images, we only select the central part with the size of $361 \times 361$ to produce new images for calculation. As shown in Figs. 12 (a)-(b) are two groups of original and central part images. Table 3 shows the detailed parameter settings of the digital manipulations. Obviously, there are a total of 82 digital manipulations. The total number of test image pairs is $82 \times 1,000 = 82,000$.

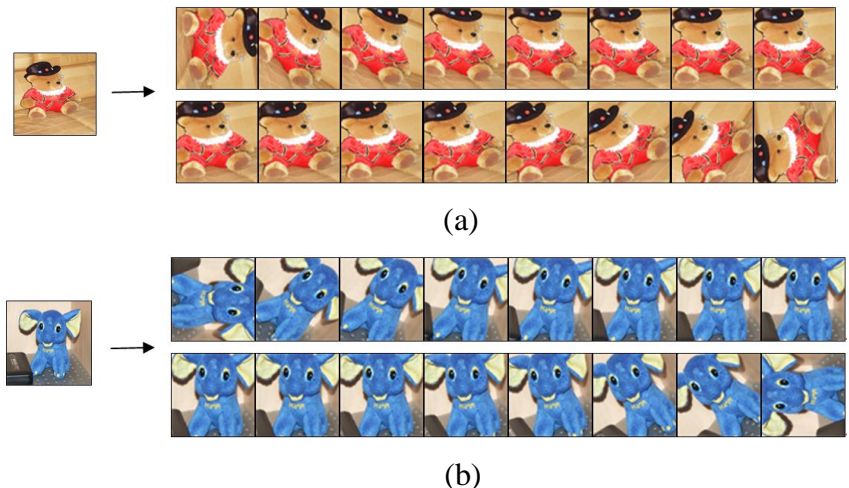

(a)

(b)

Fig. 12 Examples of the original images and central part images.

Table 3 Eight content-preserving manipulations and corresponding parameter Settings

| Manipulations | Parameter settings | Values | Number |
| --- | --- | --- | --- |
| Scaling | Scaling ratio $S_r$ | 0.8, 0.85, …, 1.2, 1.25 | 10 |
| Salt & pepper noise | Noise density $N_d$ | 0.001, 0.002, ..., 0.01 | 10 |
| JPEG compression | Quality factor $Q_f$ | 55, 60, …, 100 | 10 |
| Gaussian filtering | Standard deviation $S_d$ | 0.1, 0.2, …, 1 | 10 |
| Circular Blurring | Circle radius $C_r$ | 0.2, 0.3, …, 1.1 | 10 |
| Motion Blurring | Number of moving pixels $N_p$ | 1, 2, ..., 10 | 10 |
| Rotation and cropping | Rotation angle $R_a$ | ±1, ±0.75, ±0.5, ±0.25 | 8 |
| Large angles rotation | Rotation angle $R_b$ | ±90, ±45, ±30, ±15, ±10, ±5, ±3 | 14 |

Due to the limitation of space, six standard color images with size of 512 × 512 are exhibited in Fig. 13, i.e., *Bear*, *Tennis*, *Dessert*, *Airplane*, *Flower* and *Elephant*. As shown in Fig. 14, the parameter values of each robust attack manipulation are shown on the abscissa, and the ordinate is the Euclidean distance between the hashes of similar image pairs. Clearly, the distances on the ordinate are generally less than 400. In addition, the Euclidean distance statistical results of 1,000 standard test images in different manipulations are listed in Table 4. Observed from Table 4, the Euclidean distances of all manipulations are less than 500, except motion blurring. The values of the Euclidean distance are generally less than 700, the minimum value is less than 3.24, and the mean value is less than 237.7. For the 82 attack manipulations on all images, the maximum Euclidean distance is 762.6. Consequently, when the threshold $\xi$ of Euclidean distance was set to 740, 99.12% similar images are correctly classified.

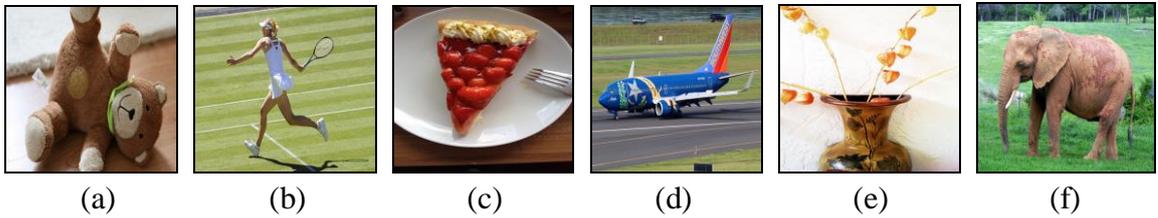

(a)  (b)  (c)  (d)  (e)  (f)

Fig. 13 Six standard test images. (a) *Bear*, (b) *Tennis*, (c) *Dessert*, (d) *Airplane*, (e) *Flower*, and (f) *Elephant*.

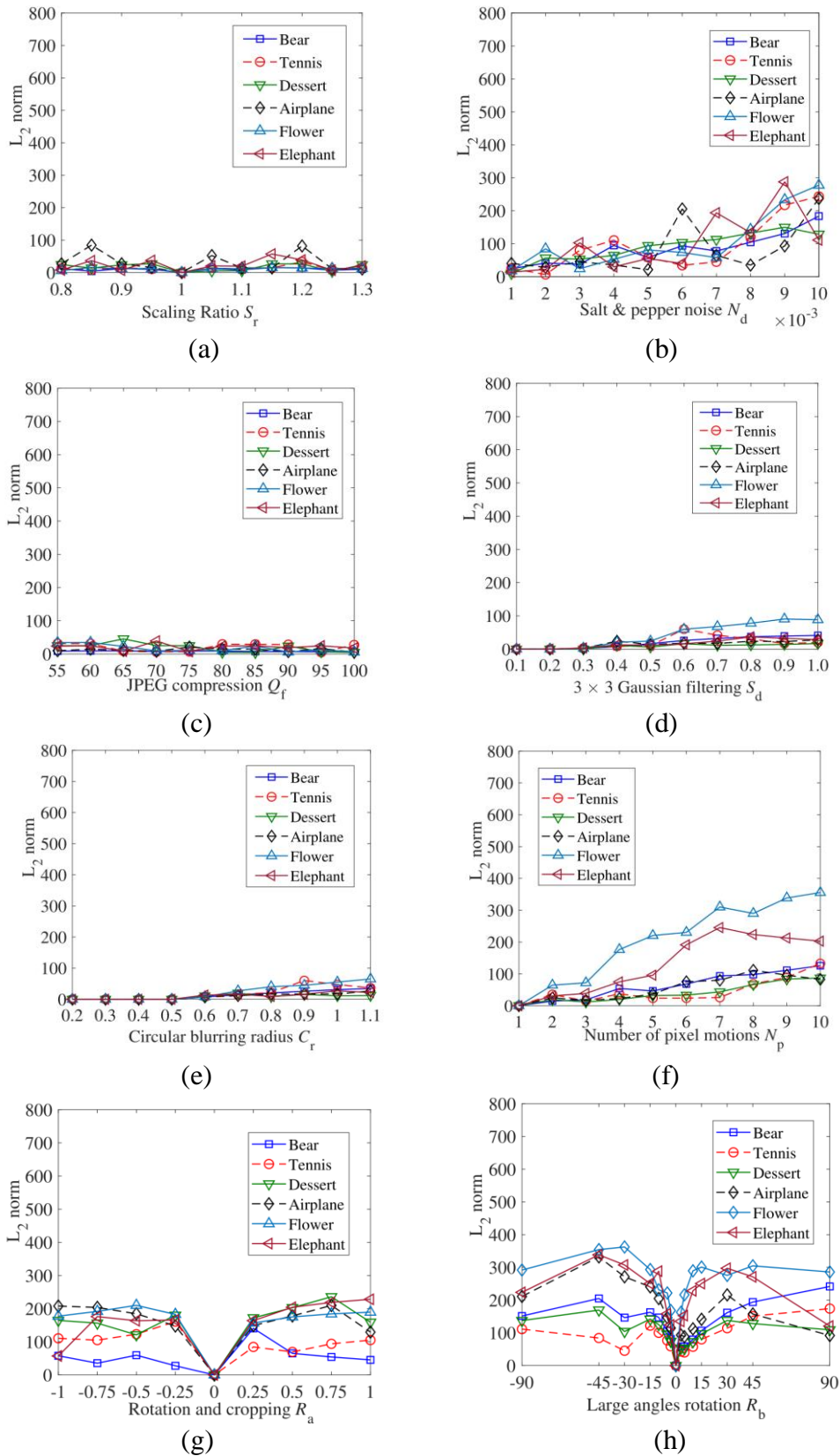

Fig. 14 Perceptual robustness representation of eight manipulations. (a) Image scaling, (b) Salt & pepper noise, (c) JPEG compression, (d) Gaussian low-pass filtering, (e) Circular blurring, (f) Motion blurring, (g) Rotation and cropping, (h) Large angles rotation.

Table 4 Statistics of Euclidean distance under different manipulations

| Manipulations | Euclidean distance | | |
|---|---|---|---|
| | Mean | Max. | Min. |
| Scaling | 54.5883 | 413.6073 | 0.0006 |
| Salt & pepper noise | 108.6966 | 487.1951 | 1.4142 |
| JPEG compression | 37.6601 | 481.4167 | 0 |
| Gaussian filtering | 34.7052 | 371.2529 | 0 |
| Circular Blurring | 52.9879 | 441.9813 | 0 |
| Motion Blurring | 237.6999 | 762.5598 | 3.2361 |
| Rotation and cropping | 132.3365 | 414.0575 | 0.0127 |
| Large angles rotation | 103.5148 | 483.0238 | 1.0001 |

### 3.3 Discriminative Capability

The 1,338 uncompressed color images in the UCID [29] database were selected in our scheme for the analysis of discriminative capability. Therefore, total number of different test images pairs is $C_{1338}^2 = 894,453$. The measurement results are shown in Fig. 15, where the x-axis and the y-axis represent the $L_2$ norm and frequency, respectively. The minimum and maximum values of hash similarity results are 354.33 and 5609.48, respectively. The mean value $\mu = 2,065.87$ and standard variation $\delta = 794.11$. With different thresholds $\xi$ are selected, the probability values of different images being misjudged as similar images are listed in Table IV. When $\xi = 740$, 0.92 % different images are incorrectly judged as visually similar images. Robustness and discrimination are the key indicators to evaluate the hash method and should be considered simultaneously. As listed in Table 5, therefore, $\xi = 740$ is selected to achieve a satisfactory effect of both perceived robustness and discrimination.

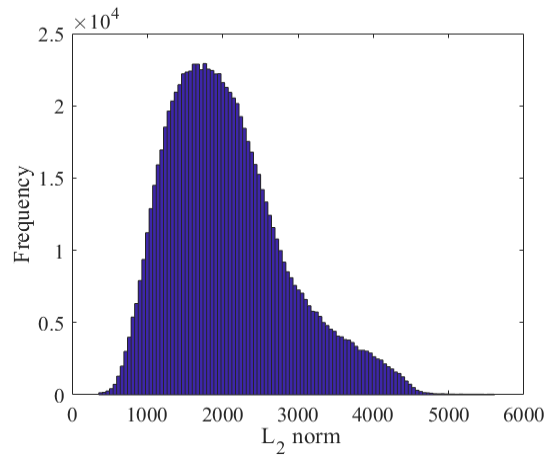

Fig. 15 Distribution of 894,453 $L_2$ norm distances between hash pairs of the 1,338 visually distinct images in UCID

Table 5 Probability of different images being judged as similar images

| Threshold $\xi$ | Probability values |
|---|---|
| 840 | 2.05 % |
| 820 | 1.71 % |
| 800 | 1.51 % |
| 780 | 1.29 % |
| 760 | 1.09 % |
| 740 | 0.92 % |
| 720 | 0.77 % |
| 700 | 0.63 % |

### 3.4 Security of Secret Keys

As shown in Fig. 16, two secret keys are utilized in the feature extraction stage and hash generation stage, which are $K_1$ and $K_2$, respectively. As shown in Fig. 16, the abscissa is 1,000 pseudo-random sequences generated keys $K_1$ and $K_2$ with wrong indexes, and the ordinate is the average of the image hashing pairs of correct and wrong keys in the normalized Euclidean distance. Almost all Euclidean distances are between 4,000 and 5,000, which is much higher than $\xi = 740$, as shown in Fig. 16. Hence, it is completely feasible to depend on the key to prevent malicious attacks.

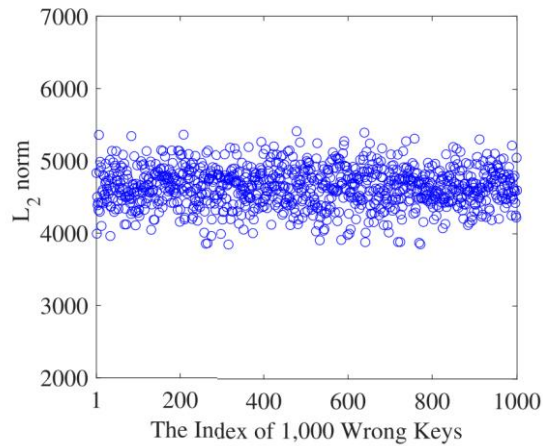

Fig. 16 Distances between the hashes of correct key and 1,000 wrong keys.

## 4. Performance Analysis and Comparison

### 4.1 Parameters Selection

*A. Color Space Analysis*

The different color spaces are suitable for different hash schemes. The choice of color space will affect many aspects. Researchers usually conduct performance analysis in YCbCr, L* a* b* or RGB color spaces. Thus, we conducted a comparative experiment on the classification performance of different color spaces to select the appropriate model. As shown in Fig. 17(a), our scheme is more suitable for YCbCr color space through comprehensive analysis of the experimental data, which curve is significantly closer to the point (0, 1) than other schemes. According to the requirements of our scheme, YCbCr color space is selected, which is more conducive to image feature extraction.

*B. Number of Ring-ribbons*

The value of $k$ affects the number of pixels in the ring-ribbon, and most pixels also contain a large amount of feature information. Here, four values of $k$ are used, i.e., $k = 61$, $k = 64$, $k = 67$, and $k = 70$. In Fig. 17(b), the ROC curve comparison amid different $k$ values. The results of ROC curves with different numbers of ring-ribbons show all well classification performance. In order to observe clearly, the details of the ROC curves near the point (0, 1) are amplified and displayed in the bottom right corner. Clearly, the number of ring-ribbon of $k = 67$ has better classification performance. It can be comprehended that a small $k$ value denotes that few feature can be identified, a large $k$ value increases the discrimination, but the perceptual robustness decreases slightly. Therefore, for size of $512 \times 512$ color images, choosing a suitable ring-ribbon number can remain the ideal balance between robustness and discrimination.

*C. Number of Corner Points*

In local color feature extraction stage, the performance of hash classification is affected by the number of corner points on each ring-ribbon outer boundary. The first $\tau$ percentage values are: 0.25, 0.4, and 0.55, respectively. By controlling other parameters remain unchanged, the ROC curves of different $\tau$ are obtained, and the experimental results are displayed in Fig. 17(c). A small $\tau$ value means only get a few points, which have little features information. Moreover, a large $\tau$ value is rich in information points. But, after the images are modified by digital manipulations, the corner point information will be affected and the perceptual robustness and discrimination will be slightly reduced. Obviously, when $\tau = 0.4$, the number of selected corner points is the most suitable.

*D. Variance Condition*

The image QD method adopts the principle of recursive decomposition to check the attribute consistency of the subintervals. And set judgment criteria to decide whether to continue iterative division. The judgment criterion, which is the selection of $V_C$ in this article, is very important for extracting image statistical characteristics. Fig. 17(d) describes the ROC curves of different $V_C$ values, when $V_C = 14$, the judgment condition is the most reasonable.

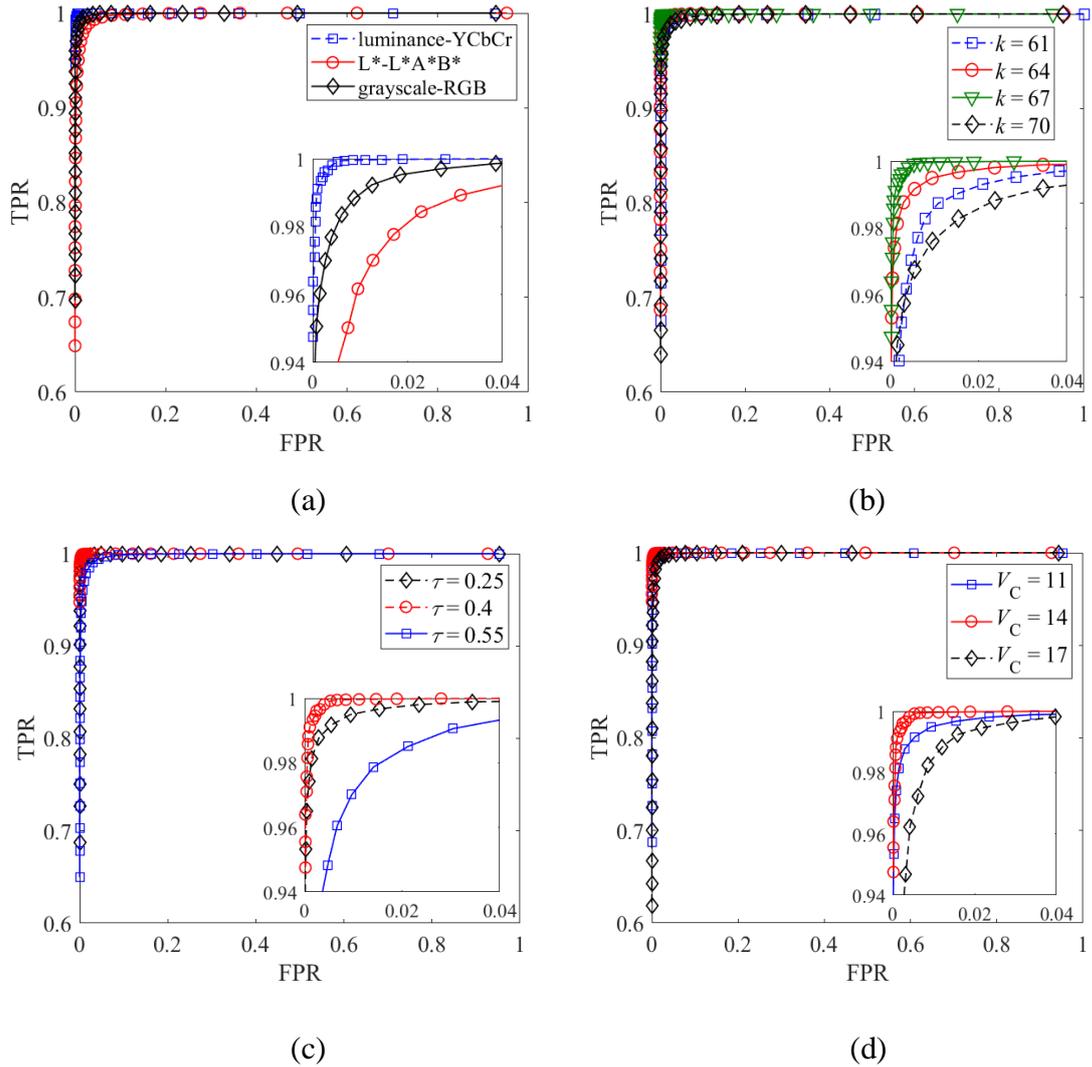

(a)                                              (b)

(c)                                              (d)

Fig. 17 Parameters selection. (a) ROC curves for different color spaces, (b) ROC curves comparison among different ring-ribbon numbers, (c) Performance comparison of different number of corners, (d) ROC curves with different variance conditions.

## 4.2 Performance Comparison

Our proposed scheme is compared with five the latest classical color image hashing schemes: Ring-NMF hashing [6], SVD-CSLBP hashing [15], BTC-CSLBP hashing [8], Ring-Entropy hashing [11] and DCP hashing [16]. These comparison schemes are not chosen arbitrarily. Since the proposed scheme is based on the ring-ribbon to resist image rotation, the contrast schemes [6], [11] are also based on the segmentation ring to extract image feature. The contrast algorithm [15], [8] are hash schemes based on LBP, which can also resist to rotation attacks. At the same time, the [16] is utilizes the most abundant corner points to extract significant structural features, which is consistent with the idea of selective sampling at the outer boundary of the ring-ribbon to extract local significant color features. Sequence similarity is measured by their corresponding methods. To make a fair comparison, we still used the same images as Section 3-B and Section 3-C to verify the classification performance. We selected some thresholds to calculate corresponding $P_{FPR}$ and $P_{TPR}$, and used ROC curve again to compare the images classification. In Fig. 18, the abscissa and ordinate of the ROC curve represent $P_{FPR}$ and $P_{TPR}$, respectively. We also compared the hash length, classification sensitivity and average time of our hash scheme and the schemes [6], [15], [8], [11], [16], see Table 6. The average time consumed in the hash generation process, the proposed scheme has vast calculation times, and the optimal scheme is the Ring-Entropy hashing [11]. Therefore, combining the contents of Fig. 18 and Table 6, it can be observed that proposed scheme has better classification performance than other comparison schemes. Although the hash length is not the shortest, but it is the most reasonable.

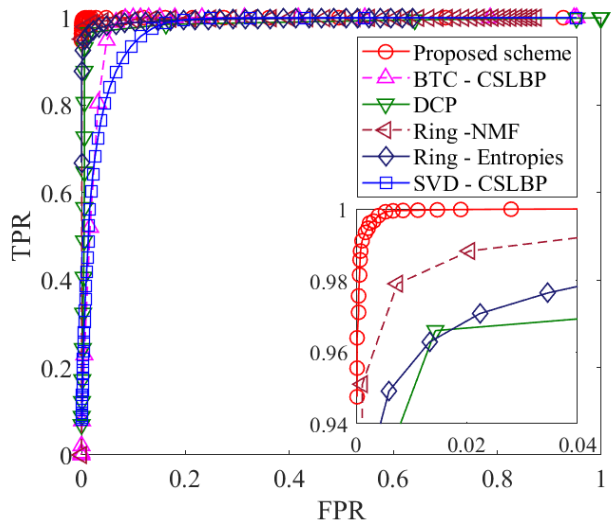

Fig. 18 Performance comparison between the proposed scheme and five comparison schemes.

Table 6 Summary of hash length and classification performance

| Schemes | Scheme [6] | Scheme [15] | Scheme [8] | Scheme [11] | Scheme [16] | Proposed scheme |
|---|---|---|---|---|---|---|
| Content sensitivity | Moderate | Moderate | Poor | Moderate | Moderate | Good |
| Hash length | 64 digits | 64 digits | 672 bits | 64 digits | 64 digits | 70 digits |
| Average time | 2.8s | 0.8304s | 0.8164s | 0.437s | 0.7618s | 3.12s |

**4.3 Application of Tampering Detection**

The Internet and various media are flooded with a large number of tampered images, which have seriously affected people's daily lives. Once the tampered picture publishes some remarks that damage the interests of the country and people, it will have an extremely serious impact on society. The Photoshop software is used to modify the color, insert or delete objects, and replace content to change the visual content of the test images. In order to verify the authentication ability of our scheme between the tampered area and the real area, this paper randomly selects six images from the Internet as the original images. The manipulates results of the image tampering methods are shown in Fig. 19, where the first and third columns are the real test images, the second and fourth columns are the tampered images. By calculating the distance between the hash pairs of the tampered images and the original test images, the calculated distances are compared with the threshold corresponding to the scheme to test the sensitivity of the visual content change. Since the Photoshop tampering method includes color modification, the scheme [11] and scheme [6] are used to compare with proposed scheme, which are all hash schemes for color images. In addition, we also compared the image authentication capability of the scheme [8] without considering the color characteristics. The similarity measures of scheme [11], scheme [6] are correlation coefficient, and scheme [8] is Hamming distance. For proposed scheme and scheme [8], when the hash distance is greater than the threshold, the image is tampered. For scheme [11] and scheme [6], when the hash distance is less than the threshold, the image is a tampered image. As listed in Table 7, in a statistical sense, compared with other schemes, the proposed scheme can effectively identify various processed images from the database.

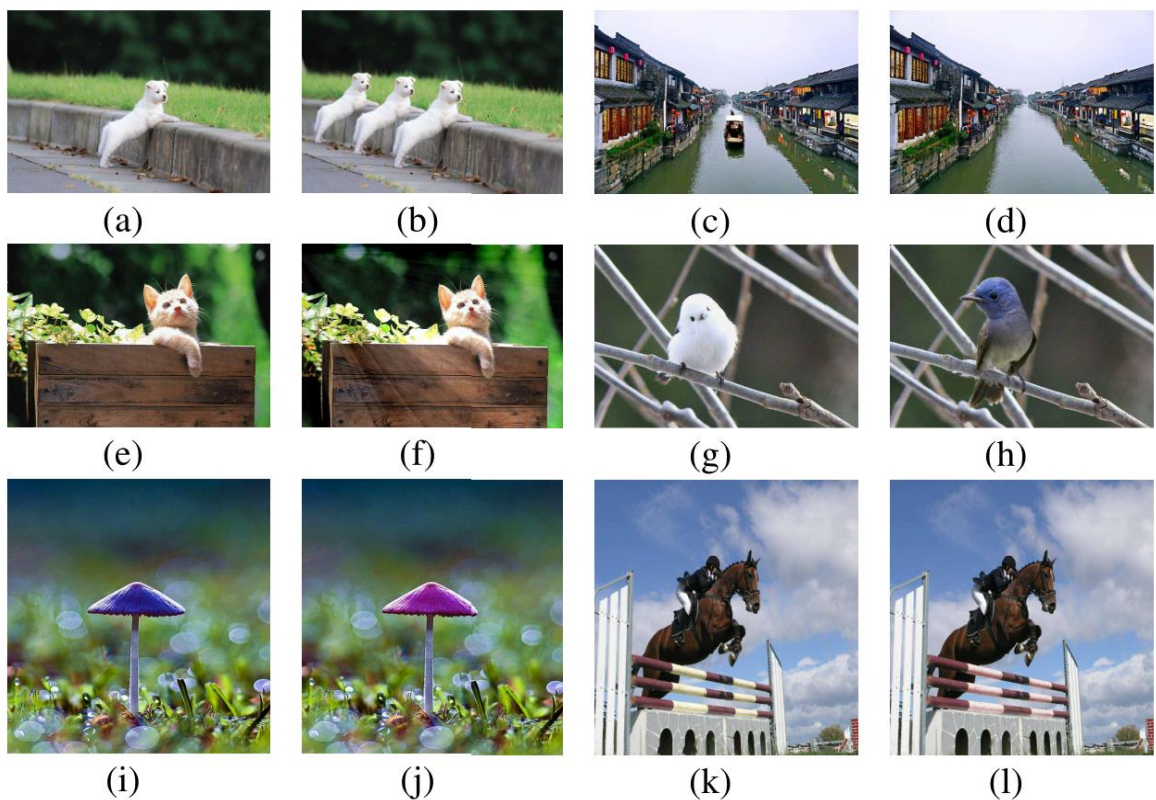

Fig. 19 Falsify the visualization result display. (a) Original image sized $960 \times 600$, (b) Tampered image : puppies inserted, (c) Original image sized $720 \times 480$, (d) Tampered image : boat deleted, (e) Original image sized $640 \times 400$, (f) Tampered image : sunshine inserted, (g) Original image sized $600 \times 395$, (h) Tampered image : bird replaced, (i) Original image sized $\times 960$, (j) Tampered image : mushroom modified, (k) Original image sized $1024 \times 800$, and (l) Tampered image : railing modified.

Table VII
THE HASHES DISTANCES BETWEEN THE ORIGINAL AND TAMPERED IMAGES[*]

| Schemes | Threshold | (a) and (b) | (c) and (d) | (e) and (f) | (g) and (h) | (i) and (j) | (k) and (l) |
|---|---|---|---|---|---|---|---|
| Ring-Entropy hashing [11] | 0.95 | 0.723 | 0.764 | 0.314 | 0.566 | <u>0.955</u> | <u>0.975</u> |
| Ring-NMF hashing [6] | 0.98 | 0.978 | <u>0.997</u> | 0.939 | 0.955 | <u>0.999</u> | 0.971 |
| BTC-CSLBP hashing [8] | 0.2 | <u>0.149</u> | <u>0.091</u> | 0.324 | 0.236 | 0.251 | <u>0.064</u> |
| Proposed hashing | 740 | 3675.259 | 1506.859 | 1233.135 | 1181.208 | 1118.279 | 866.209 |

*The thresholds in the second column are the best values in the corresponding papers. For cheme [11] and [6], the correlation coefficient is used to measure the hash similarity, the Hamming distance is utilize to measure the hash similarity of the algorithm [8], and our scheme utilize the Euclidean distance as similarity index. For the correlation coefficient method, if the value is greater than the threshold, the tamper authentication fails; nevertheless, for the Hamming distance and Euclidean distance, if the value is less than the threshold, the tamper authentication fails. Numbers marked with underscores, which indicate errors in tamper detection.

## 5. Conclusions

In this study, a scheme of combining QD and CVA is proposed. To ensure the robustness of rotation manipulation, feature extraction is performed based on ring-ribbon. The test image is normalized to eliminate interference. The QD is used to describe the texture information of image pixels, which fully reflects the perceptual content of the region within the ring- ribbon. In the color feature extraction stage, the Euclidean distance is replaced by CVA. In addition, global texture features and color features are extracted. Changing the usual sequence generation mode, a novel CCA feature fusion method is proposed to strengthen the relevance of each feature vector. During the discussion of parameters selection and performance analysis, we used ROC curve to compare the select parameters, such as different color spaces, number of ring-ribbon, and number of significant corner points. Our hash scheme is compared with the other five schemes in terms of hash length and classification performance. The proposed scheme can be used for content-based authentication and copy detection, which are well proved by the ROC curves. Experimental results show that our scheme has better performance than previous schemes.

ACKNOWLEDGEMENTS

This work was supported by the National Natural Science Foundation of China (61672354, 61702332). The authors would like to thank the anonymous reviewers for their valuable suggestions.

# References

[1] Z. Tang, L. Huang, F. Yang, Y. Dai, and X. Zhang, "Robust image hashing via colour vector angles and discrete wavelet transform," *IET Image Processing*, vol. 8, no. 3, pp.142–149, 2014.

[2] Y. Ou and K. H. Rhee, "A key-dependent secure image hashing scheme by using Radon transform," *Intelligent Signal Processing and Communication Systems*, vol. 61, no. 5, pp. 595–598, 2009.

[3] A. Swaminathan, Y. Mao, and M. Wu, "Robust and secure image hashing," I*EEE Transactions on Information Forensics and Security*, vol. 1, no. 2, pp. 215–230, 2006.

[4] Z. Tang, F. Yang, L. Huang, and X. Zhang, "Robust image hashing with dominant DCT coefficients," *Optik*, vol. 125, no. 18, pp. 5102–5107, 2014.

[5] Z. Tang, H. Lao, X. Zhang, and K. Liu, "Robust image hashing via DCT and LLE," *Computer & Security*, vol. 62, pp. 133–148, 2016.

[6] V. Monga and M. K. Mhcak, "Robust and secure image hashing via non-negative matrix factorizations," *IEEE Transactions on Information Forensics and Security*, vol. 2, no. 3, pp. 376–390, 2007.

[7] K. M. Hosny, Y. M. Khedr, W. I. Khedr and E. R. Mohamed, "Robust image hashing using exact Gaussian-Hermite moments," *IET Image Processing*, vol. 12, no. 12, pp. 2178–2185, 2018.

[8] M. Hu, K. S. Ng, P. Chen, Y. Hsiao, and C. Li, "Local binary pattern circuit generator with adjustable parameters for feature extraction," *IEEE Transactions on Intelligent Transportation Systems*, vol. 19, no. 8, pp. 2582–2591, 2017.

[9] Z. Tang, L. Chen, X. Zhang, and S. Zhang, "Robust image hashing with tensor decomposition," *IEEE Transactions on Knowledge and Data Engineering*, vol. 31, no. 3, pp. 549–560, 2019.

[10] C. Qin, X. Chen, D. Ye, J. Wang, and X. Sun, "A novel image hashing scheme with perceptual robustness using block truncation coding," *Information Sciences*, vols. 361–362, pp. 84–99, 2016.

[11] H. Hamid, F. Ahmed, and J. Ahmad, "Robust Image Hashing Scheme using Laplacian Pyramids," *Computers & Electrical Engineering*, vol. 84, 2020.

[12] Z. Tang, X. Zhang, L. Huang, and Y. Dai, "Robust image hashing using ring-based entropies," *Signal Processing*, vol. 93, no. 7, pp. 2061–2069, 2013.

[13] S. Liu and Z. Huang, "Efficient image hashing with geometric invariant vector distance for copy detection," *ACM Transactions on Multimedia Computing, Communications, and Applications*, vol. 15, no. 4, pp. 1–22, 2019.

[14] L. N. Vadlamudi, R. P. Vaddella, and V. Devara, "Robust image hashing using SIFT feature points and DWT approximation coefficients," *ICT Express*, vol. 4, no. 3, pp. 154–159, 2018.

[15] M. Sajjad, I. U. Haq, J. Lloret, W. Ding, and K. Muhammad, "Robust Image Hashing Based Efficient Authentication for Smart Industrial Environment," *IEEE Transactions on Industrial Informatics*, vol. 15, no. 12, pp. 6541–6550, 2019.

[16] R. Davarzani, S. Mozaffari, and K. Yaghmaie, "Perceptual image hashing using center-symmetric local binary patterns," *Multimedia Tools and Applications*, vol. 75, no. 8, pp. 4639–4667, 2016.

[17] C. Qin, X. Chen, X. Luo, X. Zhang, and X. Sun, "Perceptual image hashing via dual-cross pattern encoding and salient structure detection," *Information Sciences*, vol. 423, pp. 284–302, 2018

[18] C. Qin, Y. Hu, H. Yao, X. Duan, and L. Gao, "Perceptual image hashing based on Weber local binary pattern and color angle representation," *IEEE Access*, vol. 7, pp. 45460–45471, 2019.

[19] C. Yan, C. Pun, and X. Yuan, Quaternion-based image hashing for adaptive tampering localization, *IEEE Transactions on Information Forensics and Security*, vol. 11, no. 12, pp. 2664–2677, 2016.

[20] S. Liu, J. Wu, L. Feng, H. Qiao, Y. Liu, W. Luo, and W. Wang, ″Perceptual uniform descriptor and ranking on manifold for image retrieval," *Information Sciences*, vol. 424, pp. 235–249, 2018.

[21] Y. Zhao and X. Yuan, "Perceptual image hashing based on color structure and intensity gradient," *IEEE Access*, vol. 8, pp. 26041–26053, 2020.

[22] X. Wang, K. Pang, and X. Zhou, "A visual model-based perceptual image hash for content authentication," *IEEE Transactions on Information Forensics and Security*, vol. 10, no. 7, pp. 1336–1349, 2015.

[23] Q. Shen and Y. Zhao, "Perceptual hashing for color image based on color opponent component and quadtree structure," *Signal Processing*, vol. 166, No. 107244, 2020.

[24] M. Paul, R. K. Karsh, and F. A. Talukdar, "Image hashing based on shape context and speeded up robust features (SURF)," *Automation, Computational and Technology Management*, pp. 464–468, 2019.

[25] X. Nie, X. Li, C. Cui, X. Xi, and Y. Yin, "Robust image fingerprinting based on feature point relationship mining," *IEEE Transactions on Information Forensics and Security*, vol. 13, no. 6, pp. 1509–1523, 2018.

[26] Z. Tang, L. Ruan, C. Qin, X. Zhang, and C. Yu, "Robust image hashing with embedding vector variance of LLE," *Digital Signal Processing*, vol. 43, pp. 17–27, 2015.

[27] T. Lin, M. Maire, S. Belongie, J. Hays, P. Perona, D. Ramanan, P. Dollr, and C. L. Zitnick, Mictosoft coco: Common objects in context, *Proceedings of European Conference on Computer Vision*, pp. 740–755, 2014.

[28] F. A. P. Petitcolas, Watermarking schemes evaluation, *IEEE Signal Processing Magazine*, vol. 17, no. 5, pp. 58–64, 2000.

[29] G. Schaefer and M. Stich, UCID-an uncompressed color image database, *Proceedings of SPIE in Storage and Retrieval Methods and Applications for Multimedia*, pp. 472–480, 2004.